\documentclass[twocolumn,10]{article}
\usepackage[utf8]{inputenc} 
\usepackage{lmodern}
\usepackage{hyperref,url,booktabs,amsfonts,nicefrac,microtype}      
\usepackage{graphicx,caption,amsmath,comment,algorithm,algorithmic}
\usepackage{epstopdf,epsfig,multicol,tcolorbox}
\newcommand{\ci}{~\cite} \newcommand{\re}{~\ref} 
\usepackage{footnote,multirow}
\makesavenoteenv{tabular}
\makesavenoteenv{table}
\usepackage{transparent}
\usepackage[font=small,labelfont=bf,tableposition=top]{caption}
\DeclareMathOperator*{\argmax}{\arg\!\max}
\newcommand{\bde}{\begin{definition}}\newcommand{\ede}{\end{definition}}
\usepackage[hmargin=1in, vmargin=1in]{geometry}
\newtcolorbox{mymathbox}[1][]{colback=white, sharp corners, #1}

\title{Topic Modeling based on Keywords and Context}
\author{Johannes Schneider, University of Liechtenstein \and Michail Vlachos, IBM Research, Zurich}

\begin{document}
	\maketitle
	\begin{abstract}
		Current topic models often suffer from discovering topics not matching human intuition, unnatural switching of topics within documents and high computational demands. We address these shortcomings by proposing a topic model and an inference algorithm based on automatically identifying characteristic keywords for topics. Keywords influence the topic assignments of nearby words. Our algorithm learns (key)word-topic scores and self-regulates the number of topics. The inference is simple and easily parallelizable. A qualitative analysis yields comparable results to those of state-of-the-art models, but with different strengths and weaknesses. Quantitative analysis using eight datasets shows gains regarding classification accuracy, PMI score, computational performance, and consistency of topic assignments within documents, while most often using fewer topics.
	\end{abstract}
	
	
	\section{Introduction} \label{sec:Intro}
	Topic modeling deals with the extraction of latent information, i.e., topics, from a collection of text documents. Classical approaches, such as probabilistic latent semantic indexing (PLSA)\ci{hof99} and more so its generalization, Latent Dirichlet Allocation (LDA)\ci{ble03} enjoy widespread popularity despite a few shortcomings. They neglect word order within documents, i.e., documents are not treated as a sequence of words but rather as a set or `bag' of words. This might be one reason for unnatural word-topic assignments within documents, where topics change after almost every word as shown in our evaluation.  Many attempts have been made to remedy the `bag-of-words' assumption (eg. \ci{gru07,hai13,wal06,yan13}), but improvement comes usually at a price, e.g., a strong increase in computational demands and often model complexity. 

	
	The second shortcoming is an unnatural assignment of word-topic probabilities. Better solutions to the LDA model (in terms of its loss function) might result in worse human interpretability of topics\ci{cha09}. For PLSA (and LDA) the frequency of a word within a topic heavily influences its probability (within the topic) whereas the frequencies of the word in other topics have a lesser impact. Thus, a common word that occurs equally frequently across all topics might have a large probability in each topic. This makes sense for models based on document ``generation'', because frequent words should be generated more often. But using our rationale that a likely word should also be typical for a topic, high scores (or probabilities) should only be assigned to words that occur in a few topics. Although there are techniques that provide a weighing of words, they either do not fully cover the above rationale and perform static weighing, e.g., TF-IDF, or they have high computational demands but yield limited improvements\ci{lee15}.\\	
	Third, a topic modeling technique should discover an appropriate number of topics for a corpus and a specific task. Hierarchical models based on LDA\ci{teh05,mim07} limit the number of topics automatically, but increase computational costs. PLSA and LDA do not tend to self-regulate the number of topics. They return as many topics as specified by a parameter. We proclaim that this parameter should be seen as an upper bound. Choosing too large a number of topics leads to over-fitting. Furthermore, it is unclear how to choose the number of topics. Manual investigation of topics also benefits from self-regulation, i.e., a reduction of the number of topics to consider.
	\\ Other concerns of topic models are the computational performance and the implementation complexity. 	
	We provide a novel, holistic model and inference algorithm that addresses all of the above mentioned shortcomings of existing models -- at least partially -- by introducing a novel topic model based on the following rationale:\\
	\noindent i) For each word in each topic, we compute a keyword score stating how likely a word belongs to the topic. Roughly speaking, the score depends on how common the word is within the topic and how common it is within other topics. Classical generative models compute a word-topic probability distribution that states how likely a word is generated given a topic. In those models, the probability of a word given a topic depends less strongly (only implicitly) on the frequency of a word in other topics.\\
	\noindent ii) We assume that topics of a word in a sequence of words are heavily influenced by words in the same sequence. Therefore, words near a word with high keyword score might be assigned to the topic of that word, even if they on their own are unlikely for the topic. Thus, the order of words has an impact in contrast to bag-of-words models.\\
	\noindent iii) The topic-document distribution depends on a simple and highly efficient summation of the keyword scores of the words within a document. Like LDA does, it has a prior for topic-document distributions.\\
	\noindent iv) An intrinsic property of our model is to limit the number of distinct topics. Redundant topics are removed explicitly.\\
	After modeling the above ideas and describing an algorithm for inference, we evaluate it on several datasets, showing improvements compared with three other methods on the PMI score, classification accuracy, performance and a new metric denoted by topic-change likelihood. This metric gives the probability that two consecutive words have different topics. From a human perspective, it seems natural that multiple consecutive words in a document should very often belong to the same topic. Models such as LDA tend to assign consecutive words to different topics.
	
	\section{Topic Keyword Model} \label{sec:tom}
	Our topic keyword model (TKM) uses ideas from the aspect model\ci{hof01} which defines a joint probability $D \times W$ (for notation see Table\re{tab:not}). The standard aspect model assumes conditional independence of words and documents given a topic: 
	
	\begin{footnotesize}
		\begin{equation} 
		\begin{aligned}		
		p(d,w):= p(d)\cdot p(w|d) \\
		p(w|d):=\sum_t p(w|t)\cdot p(t|d) \label{eq:amod}	
		\end{aligned}	
		\vspace{-5pt}
		\end{equation} 	
	\end{footnotesize}
	
	Our model (Equations \ref{eq:m1} -- \ref{eq:m3}) maintains the core ideas of the aspect model, but we account for context by taking the position $i$ of a word into account and we use a keyword score $f(w,t)$. Taking the context of a word into account implies that if a word occurs multiple times in the same document but with different nearby words, each occurrence of the word might have a different probability. To express context, we include the index $i$, denoting position of the $i$th word in the document (see Equation \ref{eq:m1}). The distribution $p(d)$ is proportional to $|d|$. We simply use a uniform distribution. We also assume a uniform distribution for $p(i|d)$, as we do not consider any position in the document as more likely (or important) than any other. In principle, one could, for example, give higher priority to initially occurring words that might correspond to a summarizing abstract of a text.\\ 
	
	\begin{mymathbox}[title=Model Equations,boxsep=1pt,left=0pt,right=1pt,top=0pt,bottom=0pt]
		\begin{small}
			\vspace{-5pt}
			\begin{align}		
			p(d,w,i)&:= p(d) \cdot p(i|d) \cdot p(w_i=w|d,i) \label{eq:m1}\\
			p(w|d,i)&:=\max_{t,j \in R_i} \{(f(w,t)+f(w_{i+j},t))\cdot p(t|d)\} \label{eq:m2}\\	
			\text{with } R_i&:=[\max(0,i-L),\min(|d|-1,i+L)] \nonumber \\ 
			p(t|d)& := \frac{(\sum_{i \in [0,|d|-1]} f(w_i,t))^{\alpha}}{\sum_t (\sum_{i \in [0,|d|-1]} f(w_i|t))^{\alpha}} \label{eq:m3}
			\end{align} 
		\end{small}
	\end{mymathbox}
	
	To compute the probability of a word at a certain position in a document (Equation \ref{eq:m2}) we use latent variables, i.e., topics, as done in the aspect model (Equations \ref{eq:amod}). Instead of the generative probability $p(w|t)$ that word $w$ occurs in topic $t$, we use a keyword score $f(w,t)$. A keyword for a topic should be somewhat frequent and also characteristic for that topic only. The keyword scoring function $f$ computes a keyword score for a particular word and topic that might depend on multiple parameters such as $p(w|t)$, $p(t|w)$ and $p(t)$, whereas the generative probability $p(w|t)$ is only based on the relative number of occurrences of a word in a topic. We shall discuss such functions in Section \ref{sec:comp}. We use the idea that a word with a high keyword score for a topic might ``enforce'' the topic onto a nearby word, even if that word is just weakly associated with the topic. For a word $w_i$ at position $i$ in the document, all words within $L$ words to the left and right, i.e., words $w_{i+j}$ with $j \in [-L,L]$, could contain a word (with high keyword score) that determines the topic assignment of $w_i$. To account for boundary conditions as the beginning and end of a document $d$, we use $j \in [\max(0,i-L),\min(|d|-1,i+L)]$. There are multiple options how a nearby keyword might impact the topic of a word. The addition of the scores $f(w_i,t)$ and $f(w_{i+j},t)$ exhibits a linear behavior that is suitable for modeling the assumption that one word might determine the topic of a nearby word, even if the other word is only weakly associated with the topic. Generative models following the bag-of-words model imply the use of multiplication of probabilities, i.e., keyword scores, which does not capture our modeling assumption: A word that is weakly associated with a topic, i.e., has a score close to zero, would have a low score for the topic even in the presence of strong keywords for the topic. Furthermore, each occurrence of a word in a document is assumed to stem from exactly one topic, which is expressed by taking the maximum in Equation \ref{eq:m2}.
	
	We compute $p(t|d)$ dependent on keyword scores $f(w,t)$ of the words in the document $d$ (Equation \ref{eq:m3}).  We model the idea of looking for keywords in a document and aggregating their score, i.e., $f(w,t)$. The parameter $\alpha$ impacts the number of topics per document. The larger $\alpha$ the more concentrated the topic-document distribution, i.e., the fewer topics per document.
	
	Essentially, these equations allow us to derive an algorithm for inference that is efficient, while avoiding overfitting and allowing to model a prior on topic concentration.  
	
	\section{Modeling Keywords} \label{sec:comp}
	Here we state how to compute the keyword score $f(w,t)$ given a word and a topic. Alternative options are discussed in the Related Work Section.
	A word obtains a high keyword score for a topic if it is assigned often to the topic \emph{and} the relative number of assignments to the topic is high compared with other topics. The first aspect relates to the frequency of the word $n(w,t)$ within the topic. The second captures how characteristic a keyword is for the topic relative to others, i.e., $p(t|w)$. If the topic-word distribution $p(t|w)$ is uniform, the word is not characteristic of any topic. If it is highly concentrated then it is. This can be captured using the inverse of the entropy $H(w)$: 
	\begin{small}
		\vspace{-6pt}
		\begin{equation} \begin{aligned}
		H(w):=-\sum_t p(t|w)\cdot \log(p(t|w))
		\vspace{-14pt}		
		\end{aligned}  \label{eq:ptw} \end{equation}
	\end{small}
	\noindent The entropy $H(w)$ is maximized for a uniform distribution, i.e., $p(w|t)=1/|T|$ giving $H(w)=\log |T|$, where $|T|$ is the number of (current) topics, i.e., initially $k$. Thus, $1/H(w)$ is a measure that increases the more concentrated the assignment of words to topics is. However, if all occurrences of a word are assigned to a single topic, the entropy is zero and the inverse infinite. Therefore, we add one in the denominator, i.e., $1/(1+H(w))$. 
	Furthermore, if the occurrences of a word $n(w)$ in the entire corpus are fewer than the number of topics $|T|$, then the word's entropy can be at most $\log n(w) < \log |T|$. Ignoring this results in a high keyword score for a rare word even if each occurrence is assigned to a different topic. Thus, we ensure that rare words are not preferred too much by using the factor $\log \min(|T|,n(w)+1)$, where the addition of one is to ensure non-zero weights for words that occur once. An optional weight parameter $\delta$ allows more or less emphasis to be put on the concentration (relative to the frequency within a topic). Overall, our concentration score is: 
	\begin{small}
		\begin{equation} \begin{aligned}
		\vspace{-2pt}
		con(w):=\Big(\dfrac{\log(\min(|T|,n(w)+1)}{1+H(w)}\Big)^{\delta}
		\vspace{-2pt}
		\end{aligned}   \end{equation} \label{sec:ent}
	\end{small}
	\noindent The second aspect of keyword scores relates to the frequency of the word within a topic. Mathematically speaking, the frequency of a word might be estimated by the probability of the word times the total number of words, i.e.  $p(w,t)\cdot \sum_{d\in D}|d|$ or, as we shall discuss in the inference simply using the assignment counts $n(w,t)$.
	Damped frequencies, e.g., $\log (1 + p(w,t)\cdot \sum_{d\in D}|d|)$, work better than using raw frequencies for inference and classification, because classification relies more on concentration, i.e., being certain that a word belongs to a topic. For humans, the words with the highest keyword score are often too specific -- they might only be familiar to experts on the topic. Therefore, we propose a second keyword distribution targeted for human understanding that puts more emphasis on frequency using raw counts.	
	\noindent Combining the word frequency and the concentration score we get a score $f(w,t)$ that prefers rather specific keywords and a second one that emphasizes more widely used (known) words $f_{hu}(w,t)$. Both can be normalized. We add a prior $\beta$ for $f(w,t)$, similar to LDA and other models, stating that a word is assumed to occur for each topic at least $\beta$ times.
			\vspace{-3pt}
		\begin{small}
		\begin{equation} \begin{aligned}
		f(w,t)&\propto \log(1+p(w,t)\cdot \sum_{d\in D}|d|+\beta)\cdot con(w)  \\ 	
		f_{hu}(w,t) &\propto (p(w,t)\cdot \sum_{d\in D}|d|)\cdot con(w) \label{eq:pwt} 
		\end{aligned}   \end{equation} 
	\end{small}	
	\vspace{-15pt}
	\section{Inference} \label{sec:inf}
	We want to find parameters that maximize the likelihood of the data, i.e.,  $\prod_d \prod_{i \in [0,|d|-1]} p(d,w,i)$. A key challenge for inference is the fairly complex model formalized in Equations (\ref{eq:m1}--\ref{eq:m3}) and (\ref{eq:pwt}). Although methods such as Gibbs sampling might be used, they would be rather inefficient. In particular, optimizations for faster inference of a Gibbs sampler, such as integrating out (collapsing) variables, are harder to apply for a complex model. To derive an efficient inference mechanism, we follow the expectation-maximization(EM) approach combined with standard probabilistic reasoning based on word-topic assignment frequencies. The general idea of EM is to perform two steps. In the E-step latent variables are estimated, i.e., the probability $p(t|w,i,d)$ of a topic given word $w$ and position $i$ in document $d$. In the M-step the loss function is maximized with respect to the parameters using the topic distribution $p(t|w,i,d)$. 
	In our model we assume that a word at some fixed position in a document can only have one topic as expressed in Equation (\ref{eq:m2}). Therefore, the topic $t(w,i,d)$ is simply the most likely topic of that word in that context, i.e., adjusting Equation (\ref{eq:m2}) accordingly we get:
			\vspace{-11pt}					
	\begin{small}

		\begin{equation} \begin{aligned}		
		&t(w,i,d) := \argmax_{t} \{(f(w,t)+f(w_{i+j},t))\cdot p(t|d)| \footnotesize{\text{$j \in R_i$}}\} \\
		\vspace{-3pt} 			
		&p(t|w,i,d) = \begin{cases}  \label{eq:siass}
		1  \qquad t_{w,i,d}=t\\
		0 \qquad t_{w,i,d}\neq t\\
		\end{cases}				
		\end{aligned}   \end{equation}
	\end{small}	
		\vspace{-12pt} 
			
	This differs from PLSA and LDA, where each word within a document is assigned a distribution typically with non-zero probabilities for all topics. 
	In the M-Step, we want to optimize parameters. Analogously to Equations (9.30) and (9.31) in\ci{bis06} we define the function $Q(\Theta,\Theta^{old})$ for the complete data log likelihood depending on parameters  $\Theta$:
	\begin{small}\begin{equation}	
		\begin{aligned}
		&\Theta^{new} = \argmax_{\Theta} Q(\Theta,\Theta^{old}) \\ &\text{ with } 	Q(\Theta,\Theta^{old}):= \sum_{d,i,t}  p(t|D,\Theta^{old}) \log  p(D,t|\Theta)  \label{eq:opt}		
		\end{aligned}   \end{equation} 
	\end{small}
	\vspace{-8pt}
	
The optimization problem in Equation (\ref{eq:opt}) might be tackled using various methods, e.g. using Lagrange multipliers. Unfortunately, simple analytical solutions based on these approaches are intractable given the complexity of the model equations (\ref{eq:m1}-\ref{eq:m3}). However, one might also look at the inference of parameters $p(w|t)$, $p(t)$ and $p(w)$ differently. Assume that we are given the assignments of words $w$ to topics $t$ for a collection of documents $D$, i.e., $n(w,t)$. Our frequentist inference approach uses the empirical distribution: The probability of a word given a topic equals the fraction of words that have been assigned to the topic of all words assigned to the topic. Under mild assumptions the maximum likelihood distribution equals the empirical distribution (see eg. 9.2.2 in\ci{bar12}): 
			\vspace{-5pt}
	\begin{small}
		\begin{equation}	
		\begin{aligned}	
		p(w|t)&:= \frac{n(w,t)}{\sum_{w} n(w,t)}		
		\end{aligned}   \end{equation} 
	\end{small}	
		Note, that $n(w,t)$ is computable by summing across the assignments from the E-step, i.e., $p(t|w,i,d)$, because each word within a context is assigned to one topic only (Equation \ref{eq:siass}):
		\begin{small}
			\begin{equation}	
			\begin{aligned}	
			n(w,t):= \sum_{i,d} p(t|w_i,i,d)
			\end{aligned}   \end{equation} 
		\end{small}
	To compute $p(t|w)$ we use Bayes' Law to obtain $p(t|w)=p(w|t)\cdot p(t)/p(w)$. Therefore, the only free parameters, we need to estimate are $p(t)$ and $p(w)$, i.e., $k+|W|$. We have $p(t)=\tfrac{\sum_w n(w,t)}{\sum_{w,t} n(w,t)}$  and $p(w)=\tfrac{\sum_t n(w,t)}{\sum_{w,t} n(w,t)}$, thus: 
	\begin{small}
		\begin{equation}	
		\begin{aligned}	
		p(t|w) &=p(w|t)\cdot \tfrac{p(t)}{p(w)} \\ &= \tfrac{n(w,t)}{\sum_{w'} n(w',t)} \cdot \tfrac{\big(\sum_{w'} n(w',t)\big)/\sum_{w',t'} n(w',t')}{\big(\sum_{t'} n(w,t')\big)/\sum_{w',t'} n(w',t')} \nonumber\\ 
		&= \tfrac{n(w,t)}{\sum_{t'} n(w,t')}\nonumber
		\end{aligned}   \end{equation} 
	\end{small}
	We also need to compute the keyword score $f(w,t)$. We use that $\sum_{d\in D} |d|=\sum_{w,t} n(w,t)$, since each word in each document is assigned to one topic.
		\begin{small}
			\begin{equation}	
			\begin{aligned}	
			&f(w,t)\propto \log(1+p(w,t)\cdot \sum_{d\in D} |d| + \beta)\cdot con(w)\nonumber \\
				&\propto \log(1+p(w|t)\cdot p(t)\cdot \sum_{w,t} n(w,t) + \beta)\cdot con(w) \nonumber\\
				&\propto \text{\footnotesize{$\log(1+\tfrac{n(w,t)}{\sum_{w} n(w,t)}\cdot \tfrac{\sum_w n(w,t)}{\sum_{w,t} n(w,t)} \cdot \sum_{w,t} n(w,t) + \beta)\cdot con(w)$}}\nonumber \\
				&\propto \log(1+n(w,t) + \beta)\cdot con(w)\nonumber \\
			\end{aligned}   \end{equation} 
		\end{small}
\vspace{-15pt}
	
	\section{Self-Regulation of Topics}\label{sec:self}
	We only keep word-topic distributions that are significantly different from each other. Redundant topics can be removed either during inference (as done in Algorithm\re{alg:TKM}) or after inference.
	To measure the difference between two word-topic distributions we use the symmetrized Kullback--Leibler divergence.
	\begin{small}
		\begin{equation} \begin{aligned}
		KL(t_i,t_j)&:=& \sum_w p(w|t_i)\cdot \log( \frac{p(w|t_i)}{p(w|t_j)})\\
		SKL(t_i,t_j)&:=&KL(t_i,t_j)+KL(t_j,t_i) \label{eq:DT}
		\end{aligned}   \end{equation} 
	\end{small}
	The set of indexes of (significantly) distinct word-topic distributions $DT$ is such that for any two topics $i,j \in DT \subseteq T$, it holds that $SKL(t_i,t_j)\geq \gamma$. 
	
	\begin{figure*}
		\noindent\begin{minipage}[b]{0.35\textwidth}	
			\small
			\setlength\tabcolsep{3pt}
			\begin{tabular}{| c | l |}
				\hline			
				Symbol & Meaning \\ \hline
				$D$ & corpus, all documents\\
				$d$ & document from $D$\\
				$|d|$ & number of words in $d$\\
				$W$ & set of unique words in $D$\\
				$w$ & word\\
				$w_i$ & $i$-th word in a document $d$ \\
				$k$ & (maximal) number of topics\\
				$T$ & set (of indexes) of topics \\ 
				& $T \subseteq [0,k-1]$\\
				$DT$ & set (of indexes) of distinct \\
				&topics in $T$\\
				$t$ & topic $t$ from $T$\\
				$t(w,i,d)$ & topic of word $w$ assuming\\
						& $i$-th word $w_i=w$ in $d$ \\
				$L$ & length of sliding wind. to 1 side\\ 
				$\alpha$, $\beta$ & topic, word prior\\
				$\delta$ & weight for word concentration  \\
				$n(w,t)$ & number of assignments of\\
				& word $w$ to topic $t$\\
				$n(w)$ & number of occurrences of \\
				& word $w$ in $D$\\  \hline
			\end{tabular}
			\captionof{table}{Notation}\label{tab:not}
		\end{minipage}\qquad 
		\vspace{-10pt}
		\begin{minipage}[b]{0.6\textwidth}
			
			\begin{algorithm}[H]
				\caption{TKM(\small{docs $D$, nTopics $k$, Priors $\alpha$, $\beta$})}  \label{alg:TKM}
				\begin{algorithmic}[1]
					\begin{small}
						\STATE $\delta:=1.5$; $L:=7$; $p(t|d):= 1/k$; $T:=[1,k]$
						\STATE $p(w|t):= 1/|W|$+noise 
						\WHILE{$p(w,t)$ ``not converged''}
						\STATE $ n(w,t):=0$  
						\FOR{$d \in D$}
						\FOR{$i=0$ to $|d|-1$}
						\STATE $R_i:=[\max(0,i-L),\min(|d|-1,i+L)]$
						\STATE $t(w_i,i,d):=\argmax_t\{\big( f(w_i,t)+f(w_j,t)\big)\cdot p(t|d)| j \in R_i\}$
						\STATE $n(w_i,t(w_i,i,d))=n(w_i,t(w_i,i,d))+1$ 
						\ENDFOR
						\ENDFOR
						\STATE $p(t|w):=\frac{n(w,t)}{\sum_{t'} n(w,t')}$ 
						\STATE $H(w):=-\sum_t p(t|w)\cdot \log(p(t|w))$
						\STATE $con(w):=\big(\frac{\log(\min(|T|,n(w)+1))}{1+H(w)}\big)^{\delta}$
						\STATE $f(w,t):= \frac{\log(1+n(w,t)+\beta)\cdot con(w)}{\sum_w \log(1+n(w,t)+\beta)\cdot con(w)}$
						\STATE $T:=DT$ \COMMENT{\small{Keep only distinct topics (Section\re{sec:self})}}			
						\STATE $p(t|d) := \frac{(\sum_{i \in [0,|d|-1]} f(w_i,t))^{\alpha}}{\sum_t (\sum_{i \in [0,|d|-1]} f(w_i|t))^{\alpha}}$
						\vspace{-3pt}
						\ENDWHILE
						\vspace{-2pt}						
						\STATE $f_{hu}(w,t):= \frac{n(w,t)\cdot con(w)}{\sum_{w'} n(w',t)\cdot con(w')}$
						\vspace{-2pt}							
					\end{small}
				\end{algorithmic}			
			\end{algorithm}
			\vspace{-4pt}				
		\end{minipage}	
				\vspace{-6pt}				
	\end{figure*}
		
	\section{Evaluation} \label{sec:eval}
	 We assessed the tendency to self-regulate the number of topics (Experiment 1) comparing with LDA and HDP\ci{wan11}. We evaluated several metrics, such as the classification accuracy, PMI score and computation time (Experiment 2) using LDA, BTM\ci{yan13} and WNTM\ci{zuo16}. Finally, topics were assessed qualitatively for one dataset (Experiment 3).	
			\begin{table}[htp]
				\centering
				\begin{small}
					\setlength\tabcolsep{1pt}
					\begin{tabular}{| l | r | r | r | r|}
						\hline		  					Dataset & Docs & Unique & \small{Avg.} & \small{Cla-}\\
						& & Words & \small{Words} &\small{sses}\\ \hline
						BookReviews&179541 & 22211&33&8\\ \hline
						WikiBig&52024 & 137155&346&11\\ \hline
						Ohsumed&23166 & 23068&99&23\\ \hline
						20Newsgroups&18625 & 37150&122&20\\ \hline
						Reuters21578&9091 & 11098&69&65\\ \hline
						CategoryReviews&5809 & 6596&60&6\\ \hline
						WebKB4Uni&4022 & 7670&136&4\\ \hline
						BrownCorpus&500 & 16514&1006&15\\ \hline
					\end{tabular}
				\end{small}
				\caption{Datasets}\label{tab:ds}
				\vspace{-8pt}
			\end{table} 
			\vspace{-14pt}
	\subsection{Algorithms, Datasets and Setup:} 
	We compared an implementation of Algorithm\re{alg:TKM} (available on Github\footnote{https://github.com/JohnTailor/tkm}) and LDA using a collapsed Gibbs sampler\ci{gri04b} in Python, BTM made available by the authors as C++ library and the author's WNTM Java implementation. For HDP we used the Python Gensim library. For all algorithms, we used the same convergence criterion, i.e., computation stopped once word-topic distributions no longer changed significantly. For all algorithms, we ran experiments with different parameters for $\alpha$ and $\beta$. We chose the best configuration focusing on classification accuracy, i.e., $\alpha= 5/k$ and $\beta=0.04$ for LDA, $\alpha= 50/k$ and $\beta=0.02$ for BTM, $\alpha= 50/k$ and $\beta=0.05$ for WNTM and, finally, $\alpha=2.5$ and $\beta=0.05$ for TKM. For HDP we just used parameters as in\ci{wan11}. To remove redundant topics (Section \ref{sec:self}) we used $\gamma:=0.25$.

	The datasets in Table\re{tab:ds} are public and most have already been used for text classification. For the distinct review datasets (from Amazon) we predicted either the product, i.e., book, based on a review or the product category. The Wiki benchmarks are based on Wikipedia categories. We performed standard preprocessing, e.g., stemming, stopwords removal and removal of words that occurred only once in the entire corpus. All experiments ran on a server with a 64bit Ubuntu system, 100 GB RAM and 24 AMD cores operating at 2.6 GHz. 

	\subsection{Experiments:}
	\underline{Experiment 1:} 
	We empirically analyzed the convergence of the number of distinct topics $|DT|$ (See Section\re{sec:self}) depending on the upper bound $k$ of the number of topics for LDA, HDP and TKM.
	
	\noindent\underline{Experiment 2:} We compared various metrics for LDA, TKM, WNTM and BTM using $k=100$. The classification accuracy was measured using a random forest with 100 trees, with 60\% of all data for training and 40\% for testing. The time to compute the word-topic and topic-document distributions on the training data, the time to compute the topic-document distribution on the test data, the number of distinct topics $|DT|$ (Equation\re{eq:DT}) and the PMI score as proposed in \cite{new10} were also compared. PMI measures the co-occurrence of words within topics relative to their co-occurrence within documents (or sequences of words) of a large external corpus, i.e., we used an English Wikipedia dump with about 4 million documents as in \cite{new10}. For each pair of words $w_i,w_j$ we calculated the fraction of documents in which both occurred. 
	\begin{small}
		\begin{equation} \begin{aligned}
		p(w_i,w_j) &=\frac{|\{d|w_i,w_j \in d, d \in D \}|}{|D|} \\
		p(w_i) &= \frac{|\{d|w_i \in d, d \in D \}|}{|D|} \\
		PMI(w_i,w_j) &= \log \frac{p(w_i,w_j)}{p(w_i) p(w_j)}  \label{eq:PMI}
		\end{aligned}   \end{equation} 
	\end{small}
	The PMI score for a topic is the median of the PMI score for all pairs of the ten highest-ranked words according to the word-topic distribution, i.e., for TKM the distribution is $p_{hu}$ as in Equation\re{eq:pwt}. The PMI score is the mean of the PMI score of all individual topics. PMI has been reported to correlate better with human judgment than other measures such as perplexity\ci{new10}.\\  
	We introduce a novel measure that captures the consistency of assignments of words to topics within documents as motivated in the end of Section \ref{sec:Intro}. It is given by the probability that two consecutive words stem from different topics. As LDA computes a distribution across all topics, we choose the most likely topic (see Equation \ref{eq:siass}). The topic change probability $ToC$ for a corpus $D$ is defined as
	
	\begin{small}
		\begin{equation} \begin{aligned}
		ToC:=\frac{\sum_{d \in D} \sum_i I_{t(i,d)\neq t(i+1,d)}}{\sum_{d \in D} |d|}  \label{eq:ToC}
		\end{aligned}   \end{equation} 
	\end{small}
	\noindent 
	The indicator variable $I$ is one if two neighboring words have different topics and zero otherwise.\smallskip\\
	\noindent\underline{Experiment 3:} The topics using $k=20$ discovered by LDA and TKM for the 20 Newsgroups dataset were assessed qualitatively.	
	
\begin{table*}[htp]		
	\begin{small}
		\setlength\tabcolsep{2pt}
		\begin{tabular} {| c | l | l | l | l || l | l| l| l || l|l| l| l ||  l|l| l| l || l|l| l| l ||l|l| l| l|}
			\hline			
			 &  \multicolumn{4}{|c|}{\small{Classification}} & \multicolumn{4}{|c|}{PMI} & \multicolumn{4}{|c|}{Distinct} & \multicolumn{4}{|c|}{\footnotesize{Topic Change }} & \multicolumn{4}{|c|}{Training} & \multicolumn{4}{|c|}{Inference} \\ 
			Data-&  \multicolumn{4}{|c|}{Accuracy} & \multicolumn{4}{|c|}{} & \multicolumn{4}{|c|}{Topics} & \multicolumn{4}{|c|}{{\footnotesize{Probability}}} & \multicolumn{4}{|c|}{Time[Min]} & \multicolumn{4}{|c|}{Time[Min]} \\ \cline{2-25}
			set &\scriptsize{TKM}&\scriptsize{BTM}&\scriptsize{LDA}&\scriptsize{WN.}&\scriptsize{TK.}&\scriptsize{BT.}&\scriptsize{LD.}&\scriptsize{WN.}&\scriptsize{TK.}&\scriptsize{BT.}&\scriptsize{LD.}&\scriptsize{WN.}&\scriptsize{T.}&\scriptsize{B.}&\scriptsize{L.}&\scriptsize{W.}&\scriptsize{TK.}&\scriptsize{BT.}&\scriptsize{LD.}&\scriptsize{WN.}&\scriptsize{TK.}&\scriptsize{BT.}&\scriptsize{LD.}&\scriptsize{WN.}\\\hline		
			
			20Ne. &\textbf{.79} &.78 &.76&.76 &\textbf{1.6} &1.1 &1.3&1.5 &\textbf{99} &99 &100&100 &\textbf{.1} &.7 &.6&.7 &\textbf{3.7} &66 &10 &22 &\textbf{.6} &2.9 &2.0 & 3 \\ \hline 
			Reut. &\textbf{.9} &.87 &.83 &.86 &\textbf{1.4} &1.3 &1.3 &1.3&\textbf{95} &100 &100 &100 &\textbf{.1} &.6 &.6 &.7&\textbf{1.1} &17 &2.3 &5 &\textbf{.3} &.9 &.6&.9 \\ \hline 
			Web. &.81 &.81 &.81&.79 &1.4 &1.3 &1.3&1.4 &\textbf{93} &100&100 &100 &\textbf{.3} &.8 &.7 &.8 &\textbf{.8} &16 &1.9&16 &\textbf{.1} &.6 &.3&.6 \\ \hline 
			Wiki. &\textbf{.95} &.93 &.93&.93 &\textbf{2.2} &1.3 &1.5&1.3 &100 &100 &100 &100 &\textbf{.1} &.6 &.5 &.6 &\textbf{20} &516 &71&120 &\textbf{2.0} &19 &14 &19 \\ \hline 
			Brow. &\textbf{.45} &.42 &.3&.42 &1.4 &1.2 &1.4 &1.3&\textbf{92} &100 &100 &100 &\textbf{.4} &.9 &.8 &.9 &\textbf{1.0} &15 &2.1 &28 &\textbf{.0} &.6 &.2 &.7 \\ \hline  
			Ohsu. &.24 &.23 &\textbf{.36}&.33 &2.3 &2.1 &\textbf{2.4} &2.3 &\textbf{99} &100 &100&100 &\textbf{.1} &.7 &.7 &.7 &\textbf{2.9} &66 &9.1&51 &\textbf{.7} &3.0 &2.0&2.8 \\ \hline 
			Cate. &\textbf{.9} &.88 &.87&.85 &1.5 &1.4 &1.4&1.5 &\textbf{90} &100 &100&100 &\textbf{.2} &.8 &.8 &.8&\textbf{.6} &9.6 &1.3&11 &\textbf{.1} &.4 &.3&.4 \\ \hline 
			Book. &.78 &\textbf{.79} &.75&.78 &1.4 &1.3 &1.3&1.4 &\textbf{73} &100 &100 &100 &\textbf{.2} &.7 &.7 &.8&\textbf{8.1} &146 &25 &81&\textbf{5.0} &11 &14&10 \\ \hline 
		\end{tabular}	
		\caption{Comparison of TKM, BTM, LDA and WNTM. `Bold' is better at a 99.5\% significance level.}\label{tab:res}
	\end{small}
	\vspace{-15pt}
\end{table*}

	\begin{figure}[htp]
		\vspace{-10pt}
		\includegraphics[width=1.03\linewidth]{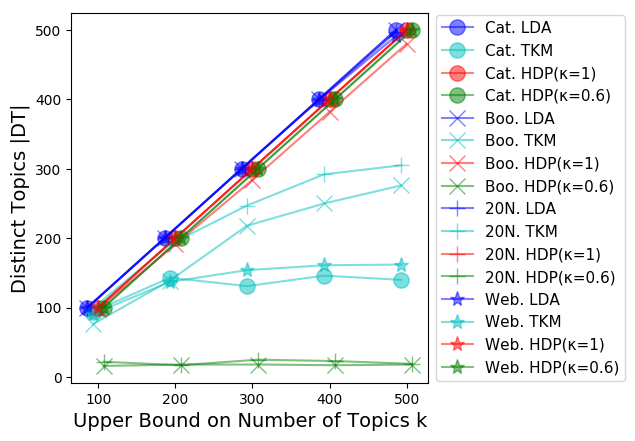}
		\vspace{-18pt}
		\caption{Distinct topics given an upper bound.}
		\label{fig:top}
		\vspace{-7pt}
	\end{figure}
		\vspace{-7pt}
	\subsection{Results:}
	\underline{Experiment 1:} As shown in Figure\re{fig:top} the number of (significantly) distinct topics $|DT|$ (see Section \ref{sec:self}) of TKM converges when the maximal number of topics $k$ increases, whereas for LDA and HDP the number of extracted topics depends (much less) on the the given upper bound.\footnote{Figure\re{fig:top} was created using small offsets on the x-axis for the sake of clarity. For visibility we only showed four datasets in the Figure.} An apparent disadvantage of HDP is that the variance in reduction seems more extreme than for TKM. HDP with $\kappa=0.6$ yields only about 30 topics for some datasets and 500 for others even though the datasets vary much less in terms of distinct words, documents and human-defined number of topics. The fact that topics are typically investigated manually and humans often assign fewer than 100 categories for each dataset (Table\re{tab:ds}) suggests that discovering more than a few hundred topics seems less suitable. HDP's convergence seems to depend more on the number of documents; that of TKM's more on the number of unique words. The number of topics $|DT|$ weakly depends on $\alpha$ -- more so for TKM. LDA showed no significant changes, even when lowering $\alpha$ to 1/500 of the recommended value of $50/k$ \cite{gri04b}. Limiting the number of discovered topics, i.e., returning very similar topics, is an intrinsic property of TKM.  A word influences the topic of nearby words directly and those of more distant words indirectly. Words with high keyword scores for a topic tend to pull other words to the same topic. \smallskip\\ 
	
	\noindent\underline{Experiment 2:}  
	Overall TKM outperforms other models on all metrics as shown in Table\re{tab:res}. We found that parameter tuning, ie. in particular of the word prior $\beta$, is essential for all methods to achieve good classification results. For classification TKM dominates or matches LDA and WNTM except for the Ohsumed dataset. The accuracy for Ohsumed is low for all techniques. It seems that there are too many terms that are shared across different categories, which seems to impact LDA the least. WNTM and LDA gave similar results, which is inline with the authors own report\ci{zuo16}. For WNTM one might expect less topic changes since it artificially creates documents based on word contexts. However, the grouping of contexts does not seem sufficient to overcome the tendency of the topic inference mechanism, ie. LDA, to group frequent words together. Overall BTM seems somewhat better than LDA and WNTM but worse than TKM for classification except for the BookReviews dataset, which is characterized by relatively short documents that seem to be better handled by BTM, which was designed for short texts. All methods can be made to do better on specific datasets, but at the price of classifying worse on others.\\ 
	PMI assesses co-occurrence and TKM shows the best results. TKM's word-topic distribution $p_{hu}$ (Equation\re{eq:pwt}) is based on weighing words by frequency and concentration. Thus, words that are frequently assigned to one topic but also to others might still get a low score. In turn, the other models focus on frequency only. It is intuitive that words that are frequent within a few topics only co-occur with higher likelihood and, thus, have a larger PMI score than words that are frequent in many topics.\\ 
	The fact that TKM tends to learn less complex models, i.e., discovers fewer topics, has been investigated in depth in Experiment 1. Less complex models are preferable when the models perform well. Indeed, TKM achieves higher classification accuracy with fewer topics. The topic change probability (Equation\re{eq:ToC}) is significantly lower for TKM, indicating more consistent assignments of words to topics. TKM tends to assign the same topic to a sequence of words, if one word in the sequence has a high keyword score for that topic. This is often the case, i.e., TKM switches topics only every 5 to 8 words. LDA and BTM switch topics after almost every other word. LDA, BTM and WNTM have a parameter that influences the number of topics per document. However, as expected for a bag-of-words model, even if a document covers few topics, the change probability tends to be high. Tuning parameters has little impact, i.e., we used $\alpha=5/k$ to reduce the number of topics (compared to the recommended value of $50/k$ \cite{gri04b}). We found that making fewer topics per document (lower $\alpha$) results in worse classification results, but still does not reduce the number of topic changes that much.\\
	The computational performance is clearly best for TKM. Speed-up ranges from a minimum factor of 2-3 for all methods for both training and inference up to a factor of 15-30 for BTM and WNTM for training. During training in our baseline implementation, the complexity per iteration is $O(L\cdot k \cdot (\sum_{d \in D} |d|))$. For LDA\ci{gri04b} it is only $O(k \cdot (\sum_{d \in D} |d|))$. The stopping criterion is the same for both algorithms. TKM requires fewer iterations and computations are simple, e.g., TKM does not sample from distributions as LDA does. Inference time of document-topic distributions is faster for TKM, as we only process a document once, whereas LDA potentially requires multiple iterations. For BTM a matrix containing all biterms is needed. This alone makes computation fairly expensive. 
	
	\begin{table*}[htp!]
		\footnotesize
		\setlength\tabcolsep{3pt}
		\begin{tabular}{| p{\textwidth}  |}\hline 
			comp.graphics, comp.os.ms-windows.misc, comp.sys.ibm.pc.hardware, comp.sys.mac.hardware, comp.windows.x, rec.autos, rec.motorcycles, rec.sport.baseball, rec.sport.hockey, sci.crypt, sci.electronics, sci.med, sci.space, misc.forsale,talk.politics.misc,talk.politics.guns, talk.politics.mideast, talk.religion.misc, alt.atheism, soc.religion.christian \\ \hline
		\end{tabular}	
		\caption{The 20 Newsgroups} \label{tab:top}
		
		\begin{tabular}{|l| l | l | l |}\hline
			&TKM Topics	& LDA Topics & Sim \\ \hline
			\scriptsize{1}&\scriptsize{ax giz tax myer presid think chz go pl ms} & \scriptsize{ax giz chz gk pl fij uy fyn ah ei} & {1.0}\\ \hline 
			\scriptsize{2}&\scriptsize{armenian turkish armenia turk turkei azerbaijan azeri} & \scriptsize{armenian turkish muslim turkei turk armenia peopl war} & {0.78}\\ \hline 
			\scriptsize{3}&\scriptsize{com bike motorcycl dod bmw ride ca write biker articl} & \scriptsize{car com bike write articl edu engin ride drive new} & {0.65}\\ \hline 
			\scriptsize{4}&\scriptsize{chip clipper encryption wire privaci nsa kei escrow} & \scriptsize{kei us com chip clipper secur encryption would system} & {0.64}\\ \hline 
			\scriptsize{5}&\scriptsize{game stephanopoulo playoff pitch espn score pitcher pt} & \scriptsize{game team plai player year edu ca win hockei season} & {0.64}\\ \hline 
			\scriptsize{6}&\scriptsize{monitor simm mhz mac card centri duo us edu modem} & \scriptsize{card drive us edu mac driver system work disk problem} & {0.64}\\ \hline 
			\scriptsize{7}&\scriptsize{gun firearm atf weapon stratu fire handgun bd amend edu} & \scriptsize{gun com edu would fire write articl peopl koresh fbi} & {0.64}\\ \hline 
			\scriptsize{8}&\scriptsize{church christ scriptur bibl koresh faith sin god cathol} & \scriptsize{god christian jesu bibl church us christ sai sin love} & {0.64}\\ \hline 
			\scriptsize{9}&\scriptsize{medic cancer drug patient doctor hiv health newslett} & \scriptsize{us medic studi diseas patient effect drug doctor food} & {0.63}\\ \hline 
			\scriptsize{10}&\scriptsize{edu israel isra arab palestinian articl write adl uci} & \scriptsize{right israel state isra peopl arab edu war write jew} & {0.59}\\ \hline 
			\scriptsize{11}&\scriptsize{space drive scsi orbit shuttl disk nasa id mission hst} & \scriptsize{space nasa gov launch orbit earth would mission moon edu} & {0.55}\\ \hline 
			\scriptsize{12}&\scriptsize{peopl us would batteri like jew right know make year} & \scriptsize{presid would state us mr think go year monei tax} & {0.49}\\ \hline 
			\scriptsize{13}&\scriptsize{imag jpeg gif format file graphic xv color pixel viewer} & \scriptsize{imag us window graphic avail system server softwar data} & {0.47}\\ \hline 
			\scriptsize{14}&\scriptsize{window mous font server microsoft driver client xterm us} & \scriptsize{file us window program imag entri format jpeg need} & {0.44}\\ \hline 
			\scriptsize{15}&\scriptsize{oil kuwait ac write rushdi edu islam uk entri contest} & \scriptsize{edu write articl com know uiuc cc would anyone cs} & {0.33}\\ \hline 
			\scriptsize{16}&\scriptsize{homosexu rutger cramer christian sexual gai msg food edu} & \scriptsize{edu peopl sai write would think com articl moral us} & {0.33}\\ \hline 
			\scriptsize{17}&\scriptsize{insur kei detector radar duke de phone system edu ripem} & \scriptsize{book post list mail new edu inform send address email} & {0.24}\\ \hline 
			\scriptsize{18}&\scriptsize{team player jesu plai water roger edu laurentian hockei} & \scriptsize{us would like write edu articl good com look get} & {0.23}\\ \hline 
			\scriptsize{19}&\scriptsize{moral atheist god widget atheism openwindow belief exist} & \scriptsize{go like get would time know us sai peopl think} & {0.14}\\ \hline 
			\scriptsize{20}&\scriptsize{car brake candida yeast engin militia vitamin steer} & \scriptsize{us com edu need power work help ca mous anyone } & {0.12}\\ \hline 
		\end{tabular}
		\caption{Topics by TKM and LDA for 20Newsgroups dataset for $k=20$ }\label{tab:qua}
		\vspace{-13pt}
	\end{table*}
	
	\noindent\underline{Experiment 3:}  Table\re{tab:qua} shows the highest ranked words per topic of LDA and TKM, i.e., using $p_{hu}(w|t)$. Using cosine similarity, topics of both methods were matched in Table\re{tab:qua}. The categories for the 20 Newsgroups dataset are shown in Table\re{tab:top}. We noted considerable variation of topics for LDA and TKM for each execution of the algorithms but overall found qualitatively comparable results. Both methods tend to find some expected topics as suggested in Table\re{tab:top} but missing or mixing others. Overall, LDA tends to give high probabilities to general words more often, e.g., `need', `problem', `go', `com', whereas TKM prefers rather specific words, e.g., `gif', `duke', `Laurentian', `BMW', `HIV'. TKM tends to mix topics using indicative words of different topics. LDA tends to find topics that are uninformative owing to the generality of the words.\\
	The first 14 topics in Table\re{tab:qua} are rather similar for both methods except Topic 12 for TKM, which makes limited sense. Topic 3 of LDA mixes autos and motorcycles. For LDA, topics 15 to 20 seem to make limited sense, as most words are common and non-characteristic of any topic. For TKM topics 15 to 20 seem to be slightly better, i.e., they tend to be more of a mix of topics with one topic often dominating. For example, topic 16 has elements of religion and sexuality, topic 18 of hockey, topic 19 of atheism and windows.x and topic 20 of autos.	
	\section{Related Work} \label{sec:rel} 
	Hofmann \cite{hof99} introduced probabilistic latent semantic analysis (PLSA) as an improvement over latent semantic analysis. Its generalization LDA\ci{ble03} adds priors with hyperparameters to sample from distributions of topics and words. LDA has been extended and varied in many ways, e.g.,\ci{liw06,new11,das15,blei10,yan13,yan15,teh05}. Whereas our model has little in common with PLSA and LDA except its rooting in the aspect model, extensions and modifications of LDA and PLSA mostly did not touch upon key generative aspects such as how document-topic distributions are determined. Most extensions with the exception of \ci{gru07,bah10,nie07,hai13,wal06,wan07,yan13} rely on the bag-of-words assumption. In contrast, we assume that a word influences topics of nearby words. In\ci{gru07} each sentence is assigned to one topic using a Markov chain to model topic-to-topic transitions after sentences. Multiple works have used bigrams for latent topic models, e.g.,\ci{bah10,nie07,hai13,wal06}.\ci{bah10,nie07,hai13} use a bigram model to improve PLSA for speech recognition. For a bigram $(w_i,w_j)$\ci{bah10,nie07} both multiply probabilities containing conditionals $w_i|w_j$.\ci{hai13} models $p(t|w_i,w_j) \propto  p(w_i|t)\cdot p(w_j|t)$. They use distanced n-grams, i.e., for a fixed distance $d$ between two words they estimate a probability distribution $p(w_i|t)$ using all word pairs at distance $d$ from the corpus. N-gram statistics and latent topic variables have been combined in\ci{wal06} and later work, e.g.,\ci{wan07,yan13}. A key underlying modeling assumption of\ci{wal06} is inferring the probability of one word given its predecessor using smoothed bigram estimators. The sparsity of short texts was the motivation in\ci{yan13} to use biterms yielding the BTM model. In the BTM model the probability of a biterm equals $p(w_i,w_j) = \sum_t p(t)\cdot p(w_i|t)\cdot p(w_j|t)$. In\ci{yan13} an increases of the time complexity by about a factor of 3 is reported together with improvements otherwise. We also consider all biterms within a window and therefore also compare against\ci{yan13}. Aside from that, there are few similarities. \\	
	Relatively little work has been conducted on limiting the number of topics. Hierarchical topic models\ci{teh05,mim07,wan11} do not require the specification of the number of topics but come with increased complexity. The hierarchial extension of LDA (HDP)\ci{wan11}, for example, has an upper bound on the number of topics and several parameters that control the number of discovered topics. For TKM, no additional parameters control the number of topics though the parameter settings influence the number of topics, in particular the upper bound on the number of topics.\\
	The word network topic model(WNTM)\ci{zuo16} is also made for short texts. For each word it creates a document by combining the contexts of the words in the original corpus. Then it uses LDA for topic inference of the created documents and a further step is needed to get topic distributions of the original corpus.\\ 
	Keyword extraction often relies on using co-occurrence data, POS tagging\ci{ros10} and external sources or TF-IDF\ci{has14}. Typically, these methods first extract key-phrases and then rank them. For example,\ci{ros10} first splits words into phrases using sentence delimiters and stop words. They compute keywords using the ratio of the frequency of a word as well as its degree, i.e., all words that are within a specific distance for any occurrence within a phrase. We do not use any of the typical preprocessing, e.g., POS tagging or phrase extraction, though this might be beneficial. We also tested the metric of\ci{ros10} and found that it gave overall slightly worse result.
	Topic modeling and keyword extraction are related. For example \ci{liu10} extracts keywords using topic-word distributions obtained from LDA. Key word extraction and clustering are also related.\ci{liu09} uses clustering as a preprocessing step to obtain keyword candidates stemming from medoids of these clusters.\ci{liu09,liu10} stick to the concept that keywords are extracted based on a preprocessing phase. In contrast, we perform a dynamic iterative approach, in which words are assigned to topics and the distribution of word assignments across topics determines the keyword score. \cite{lee15} uses term-weighing to enhance topic modeling, i.e., LDA. Their keyword score corresponds to the variance of the word-topic distribution. They do not state a thorough derivation of their Gibbs sampler, i.e., an explicit integration of Equation (4) in \cite{lee15} to get (6) and (7). Their classification performance reported on the 20 Newsgroups dataset for their best algorithm is significantly lower than the performance we observed for LDA (as well as that of our algorithms).\\
	Word embeddings and topic modeling have also been combined\ci{das15,lis16}. Our model might also benefit from word embeddings.
	\section{Conclusions} \label{sec:lim}
	We presented a novel topic model based on keywords. Both the model and its inference are simple and performant. Our experiments report improvements to existing techniques across a number of metrics, thus suggesting that the proposed technique aligns closer with human intuition about topics and keywords.
	
	\bibliographystyle{abbrv}
	\begin{footnotesize}
		\footnotesize
		\vspace{-.3\baselineskip}
		\bibliography{refs}
	\end{footnotesize}

\end{document}